\documentclass[conference]{IEEEtran}
\IEEEoverridecommandlockouts
\usepackage{cite}

\usepackage{amsmath,amssymb,amsfonts}
\usepackage{graphicx}
\usepackage{textcomp}

\usepackage{verbatim}
\usepackage{algorithm}
\usepackage{algpseudocode}
\usepackage{mathtools}
\usepackage[T1]{fontenc}
\usepackage[utf8]{inputenc}
\usepackage{lipsum}
\usepackage[table]{xcolor}
\usepackage{tabularx}
\usepackage{caption}
\usepackage{soul}
\usepackage[export]{adjustbox}
\usepackage{setspace}

\def\BibTeX{{\rm B\kern-.05em{\sc i\kern-.025em b}\kern-.08em
    T\kern-.1667em\lower.7ex\hbox{E}\kern-.125emX}}

\renewcommand{\baselinestretch}{0.99}
    
\begin{document}

\title{A Federated Learning Aggregation Algorithm for Pervasive Computing: Evaluation and Comparison
\thanks{This work has been partially supported by MIAI@Grenoble Alpes (ANR-19-P3IA-0003).}
}

\author{\IEEEauthorblockN{ Sannara EK}
\IEEEauthorblockA{Univ. Grenoble Alpes,\\ CNRS, Grenoble INP\\ LIG F-38000,\\ Grenoble, France \\
sannara.ek@gmail.com}
\and
\IEEEauthorblockN{François PORTET}
\IEEEauthorblockA{Univ. Grenoble Alpes,\\ CNRS, Grenoble INP\\ LIG F-38000,\\ Grenoble, France \\
francois.portet@imag.fr}
\and
\IEEEauthorblockN{Philippe LALANDA}
\IEEEauthorblockA{Univ. Grenoble Alpes,\\ CNRS, Grenoble INP\\ LIG F-38000,\\ Grenoble, France \\
philippe.lalanda@imag.fr}
\and
\IEEEauthorblockN{German VEGA}
\IEEEauthorblockA{Univ. Grenoble Alpes,\\ CNRS, Grenoble INP\\ LIG F-38000,\\ Grenoble, France \\
german.vega@imag.fr}
}
\maketitle

\begin{abstract}

Pervasive computing promotes the installation of connected devices in our living spaces in order to provide services. Two major developments have gained significant momentum recently: an advanced use of edge resources and the integration of machine learning techniques for engineering applications. This evolution raises major challenges, in particular related to the appropriate distribution of computing elements along an edge-to-cloud continuum. About this, Federated Learning has been recently proposed for distributed model training in the edge. The principle of this approach is to aggregate models learned on distributed clients in order to obtain a new, more general model. The resulting model is then redistributed to clients for further training.
To date, the most popular federated learning algorithm uses coordinate-wise averaging of the model parameters for aggregation. However, it has been shown that this method is not adapted in heterogeneous environments where data is not identically and independently distributed (non-iid). This corresponds directly to some pervasive computing scenarios where heterogeneity of devices and users challenges machine learning with the double objective of generalization and personalization. 
In this paper, we propose a novel aggregation algorithm, termed FedDist, which is able to modify its model architecture (here, deep neural network) by identifying dissimilarities between specific neurons amongst the clients. This permits to account for clients' specificity without impairing generalization. Furthermore, we define a complete method to evaluate federated learning in a realistic way taking generalization and personalization into account. 

Using this method, FedDist is extensively tested and compared with three state-of-the-art federated learning algorithms on the pervasive domain of Human Activity Recognition with smartphones.

\end{abstract}


\section{Introduction}

Pervasive computing promotes the integration of smart devices in our living spaces to develop services \cite{7488250,pervasiveTrend}. These devices, including sensors, actuators, and computing capabilities, enable the development of various services providing assistance to people, automating the management of infrastructures, or driving industrial processes through data analytics, etc. Many such services are already in place and take on considerable importance in numerous fields and organizations. We are now seeing the emergence of smarter services based on Machine Learning (ML) techniques. For example, such services are already being used to predict parts' wear rate in smart factories or vehicles \cite{wearRate}. 

The goal of machine learning (ML) is to train an algorithm to automatically make a decision, like a prediction or a classification, by identifying patterns that may be hidden within massive data sets whose exact nature is unknown and therefore cannot be programmed explicitly. Such systems have been tremendously successful in fields like computer vision, natural language processing, or decision making. It is then not surprising that today, there is an increasing demand to apply ML techniques in pervasive domains where traditional solutions cannot be used because of the lack of modeling tools and excessive algorithmic complexity.

Bringing such services into production nevertheless raises several problems. Current implementations are actually based on distributed architectures where models are computed in the cloud, based on data collected by physical devices. These relatively generic models are then deployed and executed on these same devices. This approach is, however, not well adapted to pervasive computing. It undergoes major limitations in terms of security (data collected in physical environments has to be sent up to the cloud), performance (communication latency is not controlled), and even costs (communications can be intensive and expensive). One interesting solution is to make more advanced use of edge resources like gateways or smartphones \cite{7488250}. The notion of the edge was mentioned in 2009 \cite{5280678} and generalized by Cisco Systems in 2014 as a new operational model. The main idea is to place computing and storage functions as close as possible to data sources.
Regarding machine learning, it comes down to offloading some learning tasks in edge resources. In doing so, data to be transferred can be reduced, and reactivity can be improved. Also, privacy can be better preserved since the nature of the exchanged information is modified. However, the lack of computing resources on edge devices and the limited amount of available data (according to learning standards) must be addressed.

Google recently proposed federated learning \cite{mcmahan2016communicationefficient,DBLP:journals/corr/abs-1902-01046,DBLP:journals/corr/KonecnyMRR16}, a new machine learning paradigm enhancing the use of edge devices. Federated learning encourages the computation of local models on edge devices and sending them to a cloud server where they are aggregated into a more generic one. The new model is redistributed to devices as a bootstrap model for the next local learning iteration.  Federated learning reduces communication costs and improves security because only models are exchanged between the edge and cloud  \cite{Lim2020}. It has immediately attracted attention as a promising paradigm that can meet the challenges of ML-based pervasive applications. Nevertheless, it has been essentially used in traditional learning fields like computer vision \cite{mcmahan2016communicationefficient,fedperr,li2019fedmd,wang2020federated} and still needs to be adapted to the specificity of the pervasive domain. In that domain, models are more likely to diverge because of data and clients heterogeneity, and because of different data volumes generated by clients. Also, clients evolve at  different paces and are connected intermittently, which involves synchronization issues.

In this paper, we propose a new aggregation algorithm, called FedDist, that has been designed to meet the specific needs of pervasive applications. We also present an extensive set of experiments, based on a well-defined evaluation method, that has been conducted to assess FedDist, as well as three other representative algorithms. Experiments were carried out in the illustrative field of Human Activity Recognition (HAR) on smartphones. This domain aims to automatically identify human physical activities, like running or walking, using sensors embedded in smartphones. HAR is well suited to our FL experiments because  activities tend to have generic patterns (an activity involves the same sequence of movements for anybody) while being highly idiosyncratic (data depends on people, devices, the way devices are carried, the environment, etc.) \cite{Stisen2015}.
Furthermore, the collected data is private and should not be sent over the network. HAR has long been addressed as a classification problem where the common approach is to process windows of data streams to extract a vector of features that, in turn, is used to feed a classifier. Many instance-based classifiers such as Bayesian Network, Random Forest, or Support Vector Machines have been used with reasonable success \cite{6181018,blachon2014}. Today, however, the most popular and effective technology is undoubtedly deep neural networks \cite{IGNATOV2018915,Cho}.

The paper is organized as follows. First, some background about federated learning is provided. In particular, the most representative aggregation algorithms 
are presented. Then, our proposed algorithm, FedDist, is detailed in Section~\ref{sec:feddist}. Then, Section~\ref{sec:eval} details the evaluation method that has been defined to evaluate and compare the different aggregation algorithms. Section~\ref{sec:expe} presents the experiments and the associated results. Finally, the paper ends with a conclusion based on our main findings and presents open perspectives and future work in Section~\ref{sec:conclu}.

\section{Background}\label{sec:sota}

\subsection{Federated Learning principles}

\begin{figure}[!htb]
\centering
\includegraphics[width=\linewidth]{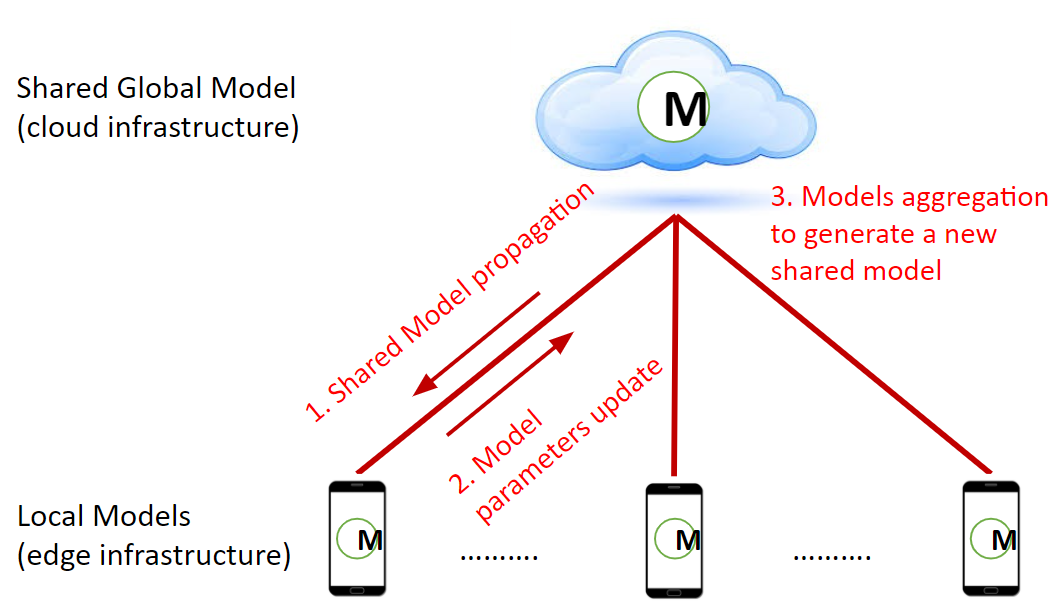}
\caption{Federated learning architecture and principles.}
\label{fig:FL_principles}
\end{figure}

As introduced before, federated learning promotes the computation of local models on edge devices and the aggregation of these models on a server to produce a new, more generic model that is sent back to all the clients. Clients start learning again from this new model and, after a while, send it to the server to do a new aggregation. This specialization/generalization cycle is repeated until a satisfactory model is obtained. A specialization/generalization cycle is usually called a communication round. After that first model convergence, more rounds may be necessary if new customers arrive or if new data is acquired. This approach is illustrated hereafter by Figure~\ref{fig:FL_principles}. 

A key point in federated learning is the way specialized models are aggregated at the server. In the case of deep learning, two families of algorithms implementing different strategies can be considered. The first strategy is to emphasize generalization. The aggregation algorithm considers local models in their entirety (all layers and neurons) and builds a new model
that potentially calls into question all layers and all the weights associated with neurons. This approach is exemplified by the FedAvg \cite{mcmahan2016communicationefficient}  and FedMA \cite{wang2020federated} algorithms, which will be detailed next. Let us also mention the FedProx algorithm \cite{li2018federated} that penalizes clients that diverge too much from the others avoid outliers. The second strategy, in contrast, focuses more on client specialization. Thus, the algorithm does not question certain parts of the local models. Specifically, only the local models' base layers are sent to the server for generalization, while the last layers are kept unchanged. This approach is exemplified by the FedPer algorithm \cite{fedperr}, which is presented subsequently.

\subsection{Federated Averaging (FedAvg)}

The FedAvg \cite{mcmahan2016communicationefficient} starts with random initialization of a neural model on a device or, more frequently, on a central server in charge of coordinating and managing model transfers with client devices. Initialization is a major step since it defines the neural model's structure in terms of layers and neurons. The resulting model is sent to clients to start local training from it. When on-device training is finished, the weights of the local models are sent to the server. Here, aggregation is done in a weighted averaging manner where clients with more data influence more significantly the newly aggregated model. The obtained model is sent back to all the clients, and another communication round can start.

FedAvg, however, has a naive form of aggregation due to its coordinate wise averaging that may lead to sub-optimal solutions. For instance, with non-Independent and Identically Distributed (non-IID) data, neurons in the same coordinate may be opted for entirely different purposes due to clients' specialization. Thus, averaging neurons that are drastically different causes decremental results. Also, the training has to go through long learning phases at each round to recover some specialization.

\subsection{Federated Learning with Personalization Layers (FedPer)}

The FedPer algorithm is similar to the FedAvg in the way it computes new weights in the aggregated models. However, it differs strongly on the parts of the model that are considered during aggregation. Precisely, clients only communicate the neural model's base layers to the server and retain the other layers. The underlying idea is that the base layers deal with representation learning and can be advantageously shared by clients through aggregation. The upper layers are more concerned with decision making, which is more specific to each client. In doing so, FedPer helps clients better handle various inputs (in the base layers) while being able to specialize in their particular data (in the upper layers). However, let us note that the aggregated model, on the server-side, is only partial and is not usable for decision-making due to the missing layers. The global approach is illustrated in figure 2. 

\begin{figure}[!htb]
\centering
\includegraphics[scale=0.08]{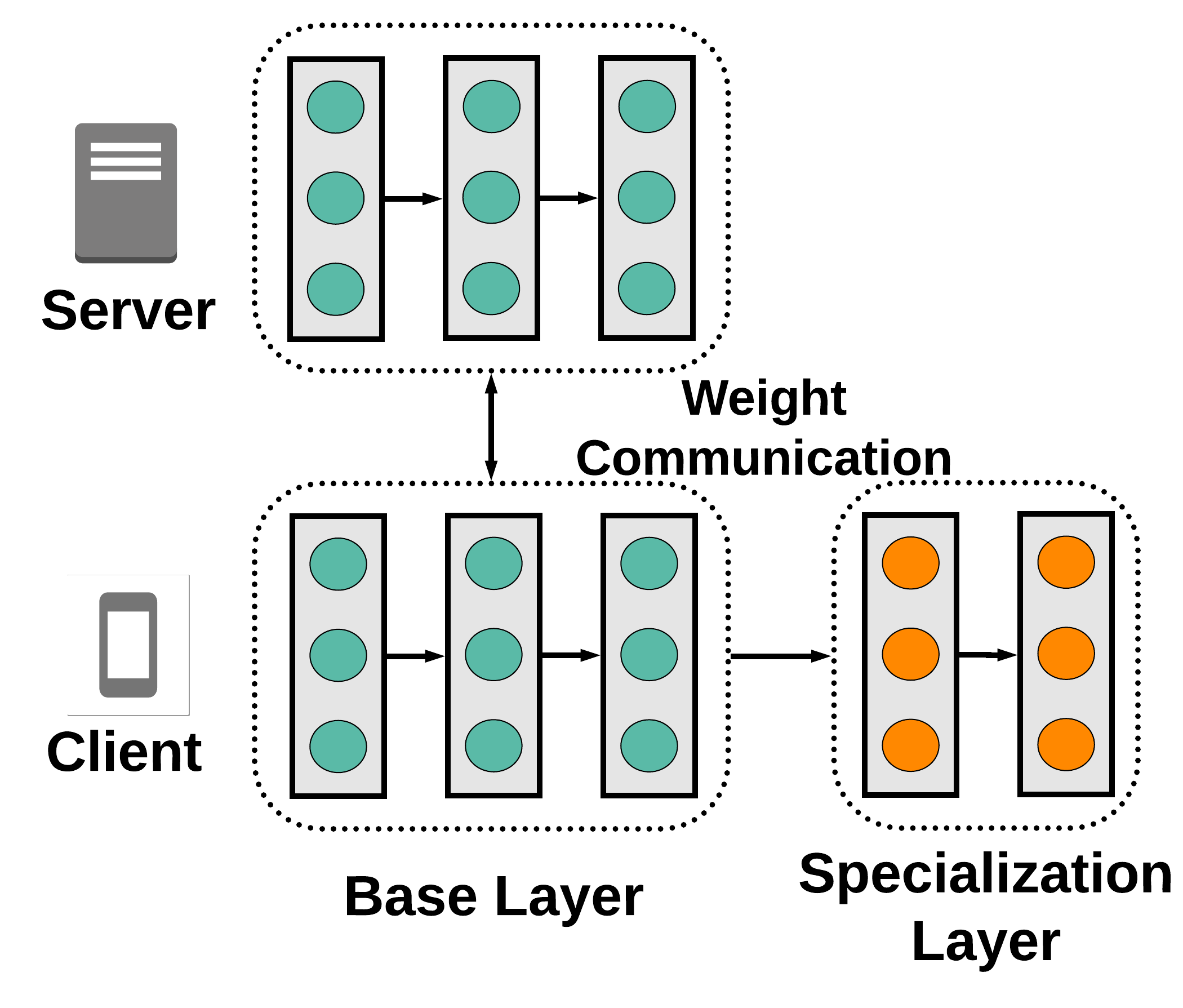}
\caption{FedPer illustration.}
\label{fig:fedPer}
\end{figure}

FedPer can be seen as an adaption of the Transfer Learning \cite{5288526} methodology into a federated learning scheme. Studies \cite{wu2020personalized} have shown that it can surpass centralized learning and FedAvg approach in the HAR field (with well-distributed datasets).

\subsection{Federated Matched Averaging (FedMA)}

The FedMA algorithm \cite{wang2020federated} is a recent algorithm that performs a more subtle model aggregation. Specifically, FedMA modifies the neural model architecture by incorporating a layer-wise aggregation process where similar neurons can be fused, and new ones can be added. This approach treats the number of nodes in a layer as a sub-problem to solve rather than a hyper-parameter to be set as an extension of \cite{pmlr-v97-yurochkin19a} to convolutional neural networks and recurrent neural networks. FedMA considers that neurons in a neural network layer are permutation invariant to perform a more intelligent aggregation. The algorithm's central intuition is that all clients can contain neurons that are similar and should be merged together. These neurons can be clustered in a non-parametric way where all neurons in the same cluster are averaged to produce a global neuron. To find which neurons can be fused, the algorithm uses a 2D permutation matrix that is computed iteratively from increasing rank layers. This matrix is calculated thanks to the Beta-Bernoulli Process - Maximum a Posteriori (BBP-MAP) \cite{pmlr-v2-thibaux07a}. The Hungarian algorithm \cite{doi:10.1002/nav.3800020109} is then used on the resulting matrix to select which local neurons can be fused and decide on which other neurons to be added. Modified layers are incrementally distributed back to clients. Note that the soft-max layer uses weighted averaging, just like FedAvg.

\begin{figure}[!htb]
\centering
\includegraphics[scale=0.08]{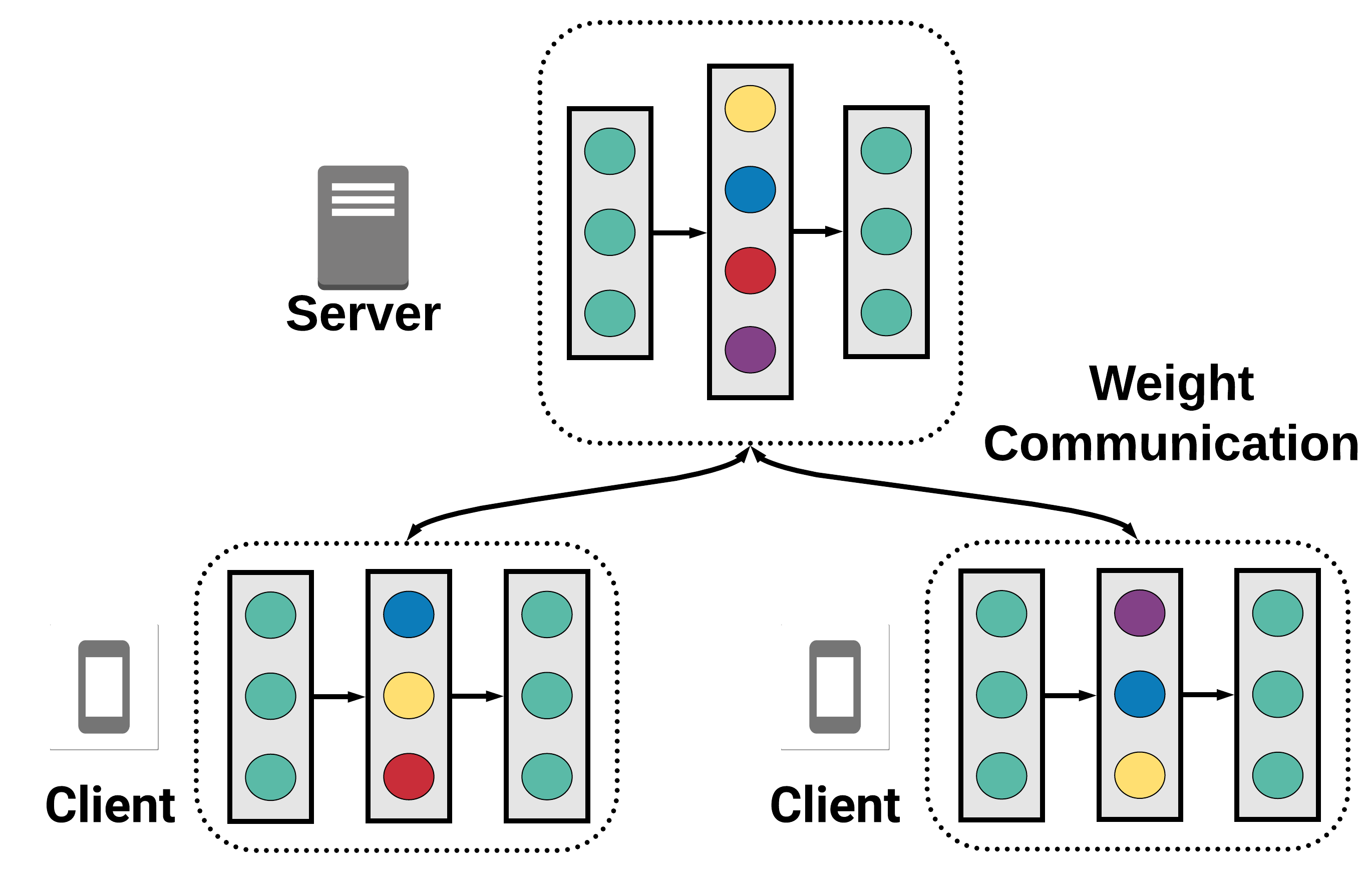}
\caption{FedMA illustration (similar neurons in same color).}
\label{fig:fedma}
\end{figure}

Experiments with deep convolutional neural networks and Long short-term memory architectures show that the FedMA algorithm outperforms FedAvg on computer vision data-sets \cite{wang2020federated}. It also appears that communications are reduced at the cost of greater complexity. Calculating the permutation matrix makes the algorithm particularly slow during the aggregation stage.

\subsection{Synthesis}

Federated learning is a very promising approach but still in its infancy. A major issue today is the lack of structured, extensive tests of the different aggregation algorithms that have been proposed. So far, these algorithms are evaluated independently on traditional learning fields like computer vision, usually on relatively homogeneous data, with no outliers or little divergence at the client-side, although federated learning has been designed to tackle unevenly distributed data\cite{konecny2016federated}. 
Also, most studies do not include the influence of the network or consider the client's behavior. 
Further studies are needed to really understand how the server and client models evolve and how well they behave regarding generalization (can a client model work correctly on data that have not been seen beforehand?) and specialization (can a client work well with data that have the same properties as the one used for training?). 

Federated learning has not been widely used in the context of pervasive computing. For instance, there are only a few works in the HAR domain \cite{wu2020personalized,8672262,8885054,NIPS2017_7029,9076082}, with missing analysis regarding the performance of global and local models on generalization and specialization with different approaches. Additionally, most tests were performed on small and pre-processed datasets collected in laboratory environments. Also, existing algorithms do not consider the specific requirements of pervasive computing, where data are usually skewed, heterogeneous, and non-IID. For instance, FedAvg is bounded to a single model shape with a fixed amount of filters or neurons to tackle an ever-growing problem (since new data and clients keep coming). Finally, contrary to centralized learning, where the model is learned to increase performance on one unique dataset, in FL the objective is to increase all individual clients' performances. To do so, FL must account for the fact that data, even though in the same domain, might evolve very differently according to the clients and would necessitate adapting the features to be induced (i.e., the representation must be learned again).  In order to account for the heterogeneity of clients, a solution must be found to adapt the global model in a way that makes it respect the peculiarity of each client. The FedDist algorithm presented below is our attempt to reach this goal.

\section{FedDist, a new federated learning algorithm}\label{sec:feddist}

In this section, we present a novel neuron matching and detail a federated learning algorithm based on a euclidean distance dissimilarity measurement. This algorithm, which includes some elements of FedAvg and FedMA, is called FedDist (Federated Distance) for its emphasis on computing distances of neurons of similar coordinates when comparing clients and server models. 

FedDist recognizes, like FedMA, that some client's models may diverge because of heterogeneous,  non-IID data. This results in neurons that cannot be matched with neurons from other models (because of weights that are too far apart). A naive coordinate averaging approach, like in FedAvg, has very negative effects. Indeed, these diverging neurons are simply erased by the averaging process, while they are actually able to deal with specific situations not encountered by other clients. FedDist also recognizes that the model's structure is relatively stable, which means that neurons with the same coordinates play a similar role. This view provides an opportunity to build on a coordinate-wise approach. 

\begin{figure}[!htb]
\centering
`\includegraphics[width=\linewidth]{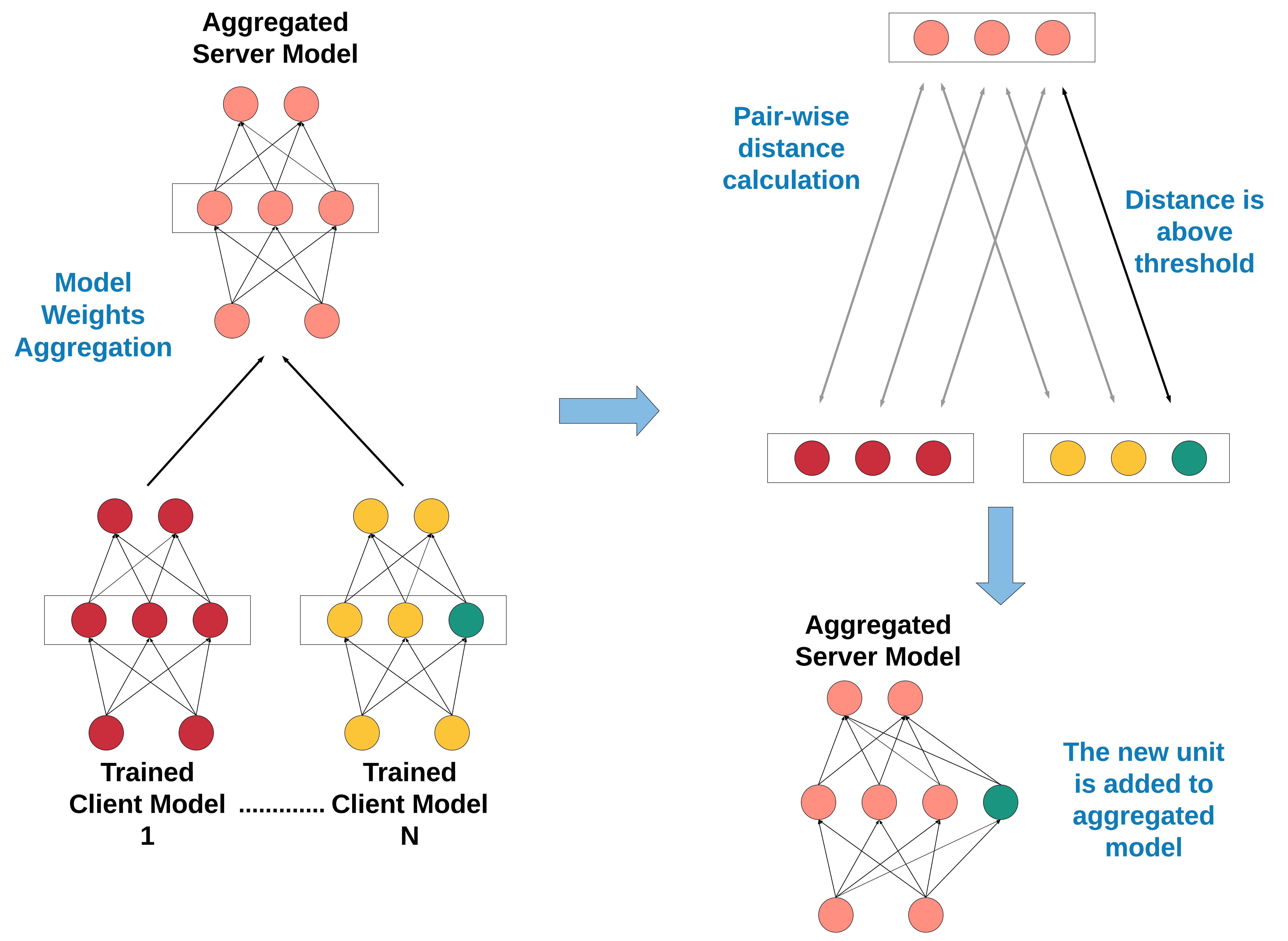}
\caption{FedDist unit generating process}
\label{fig:fedDistProcess}
\end{figure}

FedDist identifies diverging neurons using euclidean distances. These neurons that are specific to certain clients are added to the aggregated model as new neurons.  We believe that this technique is particularly suited for tackling sparse data, where specific features are only found in a small subset of clients or data points. Thus, this new neuron adding scheme can lead to larger models that are able to generalize better. As new neurons are added to a layer, a layer-wise training round is added in order to allow the neurons in the next layers to adjust to the new incoming neurons and weights. To do so, the layer with the new neuron and those below are frozen, and the subsequent layers are trained. This introduces intermediate communication rounds: a full communication round is finished when all layers have been treated, and a new aggregated model is computed. This process is summarized in Figure~\ref{fig:fedDistProcess}. First, a global server model is computed from all clients (left part of the figure). Outliers are identified using the euclidean distance (upper-right) and added to the aggregated model (lower-right) sent back to the client. 

The detailed algorithm is presented hereafter (Algorithm~\ref{fedDistAlgo}). It starts by distributing the server model ($w_t$) to all clients ($w_{tk}$). These clients then commit to training locally their own model, where we denote the process as the function $ClientUpdate\_start$, and send it to the server where a weighted averaging is performed similarly to FedAvg.  A pairwise euclidean distance (see equation~\ref{eq:l2} in section \ref{sec:eval} for details) is iteratively calculated for each unit in a layer between the client models and the aggregated server model to generate a cost-distance matrix ($\prod$).

\begin{algorithm}
\caption{Federated Distance (FedDist)}
\label{fedDistAlgo}
\begin{algorithmic}[1]
\Require L = ModelLayerCount, T = CommunicationRound, K = ClientCount  
\State initialize $\emph{w\textsubscript{1} on server}$
\For{each communication round $t = 1, 2, \cdots T$}
    \For{$k = 1, 2, \cdots K$}
        \State $w_{tk} = w_{t}$
        \State $w_{tk} \gets$ k.ClientUpdate\_start($w_{tk}$)
    \EndFor
    \State $w_t \gets$  $\sum_{k=1}^{K}$  $\dfrac{n_k}{n}$  $w_{tk}$
    \vskip 3pt
    \For{each layer $l = 1, 2, \cdots L-1$}
        \For{each client $k = 1, 2, \cdots K$}
            \State $\prod_{t}^l \Leftarrow calculatePairWiseDistance(w_{t}^l,w_{tk}^l)$
        \EndFor
        \State $\mu^l,\sigma^l \Leftarrow calculateMean\&stdOfNeuron(\prod_{t}^l)$
        \State $newNeuron = False$
        \For{each neuron distance $d = 1, 2, \cdots D$ in $\prod_{t}^l$}
        \vskip 2pt
            \State $threshold = 3*\sigma_{d}^l + \mu_d^l + penaltyFunc(t)$
            \If{$mean(d) > threshold $}
                \State $appendNeuronTo(w_{t}^l)$
                \State $newNeuron = True$
            \EndIf
        \EndFor
        \If{$newNeuron$}
            \For{each client $k = 1, 2, \cdots K$}
                \State $w_{tk}^l = w_{t}^l$
                                        \Comment{Freeze layers $l$ and below}
                \vskip 2pt
                \State $w_{tk}^l \gets$ k.ClientUpdate\_start($w_{tk}^{l+1} ,...., w_{tk}^{L}$)
                \vskip 2pt
                \State $w_{t} \gets$  $\sum_{k=1}^{K}$  $\dfrac{n_k}{n}$  $w_{tk}^l$
            \EndFor
        \EndIf
    \EndFor
    \State $w_{t+1} \gets$  $w_{t}$
\EndFor

\end{algorithmic}
\end{algorithm}

Then, the mean ($\mu$) and standard deviation ($\sigma$) of the euclidean distance for every neuron are calculated. The average provides information about the direction taken by most clients. With the normal distribution property, the values less than one standard deviation away from the mean holds for 68.27\% of the set, while two standard deviations holds for 95.45\% and three standard deviations for 99.73\%. Thus, by using this distribution property, we define a threshold as below:
\newline

$threshold = 3*\sigma + \mu + penaltyFunc(comRound)$
\newline

A penalty function has also been implemented to raise the threshold as training continues to prevent the never-ending addition of new neurons. If a neuron in any of the clients holds an individual distance above the threshold, it is then added to the server model.  The process is performed layer-wise. At each communication round, it is performed on the first layer. Then the first layer is updated and frozen at the client-side. The model is then retrained on the client-side (all already treated layers frozen) to allow the next layers to adjust and adapt to any newly added neuron weights. In this stage, the client model only needs to send back unfrozen layers of the model to reduce communication overhead. 

If we consider all layers of equal size, the average communication overhead with respect to FedAvg is thus $\frac{(L-1)}{2}*FedAvg_{cost}$ since for $L$ layers, the process is repeated $L$ times with a decrementing number of layers. In practice, the cost is much lower since layer-wise training is skipped for layers where no unit has been added in the previous layer. Additionally, if all client models become fully saturated with no new neuron added, then layer-wise training is no longer needed, and the learning becomes equivalent to FedAvg.

\section{Evaluation Method}\label{sec:eval}

A proper reproducible evaluation of FL algorithms requires a clear definition of the evaluation method and baseline systems. This section describes the datasets considered in the study, the baseline HAR models, the evaluation strategy, and the evaluation metrics. 

\subsection{HAR Task and Datasets} 
In this paper, we favor reproducibility, heterogeneity, and realistic situations. Hence, the ideal dataset(s) should be: freely accessible, acquired in real-life environments with several participants and devices, include high-class imbalance, and carefully annotated. Furthermore, since FL implies several local learning phases, the dataset should be large enough to simulate an extended period of time. 

Despite HAR Task being well investigated, attempts to benchmark it on smartphones are only recent. As reported in a recent survey  \cite{Sussex_challenge}, a large number of datasets acquired from smartphones, worn in different ways, with various sensors and sampling frequency, make it difficult to reach a uniformity in tasks, sensors, protocols, time windows, etc. 

Furthermore, some datasets are very imbalanced because activity distributions among classes are very different. HAR is thus a perfect fit for testing FL in realistic scenarios. 

In the HAR domain, we have identified more than 20 datasets. A well-known example is the UCI dataset \cite{Anguita2013APD} which has been widely used as a benchmark in the domain. However, this dataset is not realistic (it was acquired in-lab following strict scenarios), and it is small in size (3.6 hours). Since the size is an important requirement of our study, we have selected the REALWORLD dataset \cite{realword}, which contains 125 hours of recorded data, including accelerometer and gyroscope readings. Data were collected from 15 subjects in 7 different device/body position configurations, using Samsung Galaxy S4 and LG G Watch R with a sampling rate of 50 Hz. The recording was performed outdoor, where the subjects were told to perform specific activities without any restraints. Eight activities have been labeled in the data: Climbing Down, Climbing Up, Laying, Sitting, Standing, Walking, Jumping, and Running. 

This dataset is in-line with a survey on HAR on smartphones \cite{overviewHar} that has shown that the optimal sampling frequency is between 20 Hz and 50Hz and that accelerometers and gyroscopes are the most adequate sensors for classification. We believe the REALWORLD dataset represents well HAR data in the wild and exhibits realistic high-class imbalance (for instance, the `standing' activity represents 14\% of the data while the `jumping' one is limited to 2\%). Furthermore, the dataset is particularly relevant for the study since each client can be modeled by a single participants' own data as one would expect in real pervasive computing scenarios.

\subsection{Baseline HAR models} 
Since FL is a meta-learning scheme, it is important to choose the classification model for the HAR task carefully. State-of-the-art approaches in HAR on smartphones have shown a broad appeal for embracing shallow neural networks. 
Table~\ref{table:UCISOTA} summarizes the performance of recent state-of-the-art models on the UCI test set, which despite its small size and variability, is a \emph{de facto} benchmark in the HAR domain. It can be seen that Convolution Neural Networks (CNNs) models are widely used for HAR due to their ability to model features from raw data. Furthermore, they present a smaller size than more complex neural network architectures, which only slightly outperform basic CNN \cite{DBLP:journals/corr/WangCHPH17}. This is an important aspect that helps reduce the costs of communication and on-device computing. It is important to note that these results are not entirely comparable since some models used handcrafted features and sometimes different learning sets.  

\begin{table}[!htb]
\centering
\caption{State-of-the-art models and accuracy on UCI.}

\resizebox{\linewidth}{!}{
\begin{tabular}{|ccc|}
\hline
 \textbf{Reference} & \textbf{Models} & \textbf{Accuracy (\%)} \\ \hline
                    Ronao and Cho, 2016 \cite{article}  & 3 * Conv + Dense layer            & 94.79                  \\ \hline
                                        Jiang and Yin, 2015 \cite{10.1145/2733373.2806333}   & 2 Conv + Dense Layer              & 95.18                  \\ \hline
                                        Ronao and Cho, 2015 \cite{inproceedings1}   & 3 * Conv + Dense Layer              & 94.79                  \\ \hline
                                         Almaslukh, 2017 \cite{Almaslukh}  & 2 * Dense Layer (SAE)            & 97.50                  \\ \hline
                                          Ignatov, 2018 \cite{IGNATOV2018915} & 1 * Conv + Dense Layer              & 96.06                  \\ \hline
                                          Anguita et al., 2013 \cite{Anguita2013APD} & SVM             & 96.37                  \\ \hline
                                           Cho and Yoon, 2018 \cite{Cho}&  DT + 2 * CNN & \textbf{97.62}                  \\ \hline
\end{tabular}
}
\label{table:UCISOTA}
\end{table}

In our study, we will then use standard CNN models that will be tuned specifically for the task.

\subsection{Evaluation strategy}\label{sec:eval_strategy}

For the sake of reproducibility, all our experiments have been performed in simulation mode (like most federated learning evaluations). However, let us note that we have also implemented our approach on real devices, i.e., Google's Pixel 2, to check the approach's feasibility. This is not reported here due to a lack of space.

As previously explained, the interest of federated learning over classical learning is the ability to merge several client models into a global one in order to improve genericity without degrading specialization. To assess these properties, we then decided to compute three different metrics for each experiment:
\begin{enumerate}
    \item \textbf{Global accuracy}. This accuracy is computed by the server with the global model aggregated from all the clients. It tells how well federated learning is able to create a general model and permits to answer the research question "\emph{Does FL bring better global performances than centralized global learning?}"
    \item \textbf{Personalization accuracy}. This accuracy is computed by the client using its local datasets. It tells how personalized the client models are and permits to answer the research question "\emph{Does FL bring better local performances with already seen data than local learning only?}"
    \item \textbf{Generalization accuracy}. This accuracy is computed by the client using the global dataset. It tells how well the client models are able to retain generalization and permits to answer the research question "\emph{Does FL bring better local performances with unfamiliar data than local learning only?}" This evaluation is crucial because it qualifies one of the main potential benefits of the federated learning approach. 

\end{enumerate}

\begin{figure}[!htb]
\centering
\includegraphics[width=\linewidth]{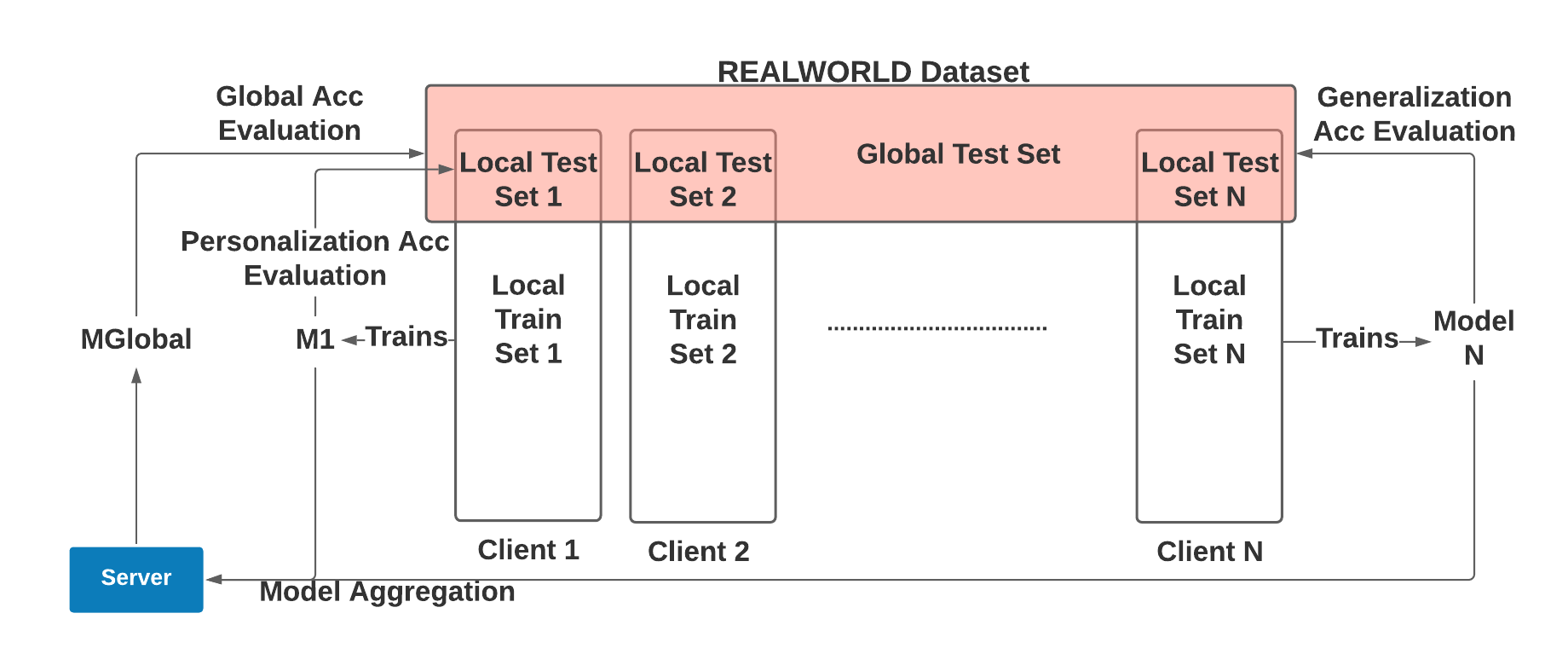}
\caption{Depiction of data partitioning/arrangement}
\label{fig:datasetBreak}
\end{figure}

Another important question of the evaluation is how to model clients. Many FL studies partition a unique dataset into several equally distributed datasets to represent the clients. This is not a realistic way of simulating heterogeneous clients. In our study, each client is represented by the record of a single identified human participant, without combining data from several participants. In that way, the FL algorithm is left dealing with a very personalized local model, which corresponds to realistic settings. Figure~\ref{fig:datasetBreak} illustrates how partitioning has been designed. Each client is represented by a set of records of one participant extracted from REALWORLD.

For each client, the dataset is partitioned into a training set and a test set. The test set is used for the local evaluation of each client. These evaluation results are then aggregated to give the \textbf{Personalization accuracy} evaluation. The concatenation of all client test sets forms the global test set, which is used to evaluate the \textbf{Global accuracy}. It is also used for the local evaluation of each client. These evaluation results are then aggregated to give the \textbf{Generalization accuracy} evaluation.

\subsection{Metrics}\label{sec:metrics}

For HAR performance metrics, we have used standard classification metrics such as accuracy, precision, and recall. From these, we have computed the F1 score as a complementary measure that gives a per-class evaluation, which is the harmonic mean of precision and recall. It is defined as below:

\begin{equation}
F_{1\_score} = \dfrac{2 \cdot Precision \cdot Recall}{Precision + Recall}
\end{equation}   

Since HAR datasets are often imbalanced (e.g., "jump" activity far less present than `` walking'' activity), we also reported the individual F1 scores' macro average. The macro F1 score is defined as:

\begin{equation}
macro\   F_{1\_score} =  \sum_{i=1} F_{1\_score} (i)  / |C|
\end{equation}

where $F_{1\_score}(i) $ is the F1 score of the $i^{th}$ class and $C$ is the set of classes. In this way, a model that performs well only with the majority classes will be penalized by the minority classes. 

While classification metrics are interesting for assessing final performances, they do not give insight into how each FL algorithm modifies the neural network layers. Indeed, after a few local-epochs of training on local devices, the client and server model weights may differ drastically. In the case of CNNs, a client filter may specialize in detecting a specific feature. After a few communication rounds, the aggregation at the server model should make the first layer of all clients identical with only the final layers still personalized by the local dataset.

To monitor this aspect, we proposed a \emph{Pair-wise Dissimilarity measure} which evaluates how different neurons are between the server and the client model. When the client models are very close, the distance between neurons of different clients should be relatively low. In this work, we used an equivalent of the L2 norm on the difference between two vectors, that is the Euclidean distance between the weight of two neurons as below:

\smallskip
\begin{equation}\label{eq:l2}
\scriptstyle {dist(N_1,N_2) = \sqrt{ (N_1^{w_1} - N_2^{w_1})^2 + \dots + (N_1^{w_K} - N_2^{w_K})^2 }}
\end{equation}
\smallskip
where $N_1$ and $N_2$ are two different neurons and $w_i$ is the $i^{th}$ of the $K$ weights of the neurons. A large distance would indicate a strong dissimilarity, while neurons that are very similar to one another should have a small distance.

Such measure is particularly useful to assess how diverging the client weights are from the server ones. This enables to identify clients that become outliers during the learning by constantly diverging from the server model. This measure is also the basis of the distant measure used in FedDist. 

\section{Experiment and Results} \label{sec:expe}

\subsection{Settings}
The four different algorithms were evaluated using the REALWORLD dataset. Each client has been simulated using data corresponding to the dataset's individual participant, leading to 15 different clients. Each client dataset was, in turn, partitioned into an 80\% -- 20\% ratio to obtain local train and test datasets. Furthermore, the local test sets were aggregated into a global test set, used to evaluate the genericity/personalization trade-off of the different algorithms.

We built and evaluated three models. In the \textbf{Global accuracy} evaluation, we tested the aggregated model against the combined global dataset. Let us note that this was not possible for the FedPer algorithm, which lacks a global model. In the \textbf{Personalization accuracy} evaluation, we tested each client model on its own local dataset. In the \textbf{Generalization accuracy} evaluation, we tested each client model on the combined global dataset. For each evaluation, we computed the accuracy, recall, precision, and F-score. 

The input data of the REALWORLD dataset were provided by the Inertial Measurement Unit (IMU) of smartphones. That is the 3-axis accelerometer data with the 3-axis gyroscope data, sampled at 50hz. As common in deep learning approaches (i.e., features are learned and not hand-crafted), no features extraction was applied to let the model build its own representation. The data was preprocessed using channel-wise z-normalization and sampled using a window-frame size of 128 with a 50 overlap of 6 channels of each axis.

All experiments were implemented using TensorFlow 2 \cite{abadi2016tensorflow} and run on a Debian 4.19.132-1 version 10 using a GPU GeForce GTX TITAN Black 6GB. We used our own implementation of FedAvg and FedPer to overcome the limitations of TensorFlow Federated (TFF) (memory size) at the time of the experiment. FedMA was adapted from the own code of the authors with minimal modifications. FedDist was entirely implemented by us. The FedAvg, FedMA, and FedPer implementations were able to reproduce the results of the original authors. 

\subsection{HAR model with Traditional Learning}

To select a state-of-the-art HAR model, we performed several experiments where we trained models without any collaborative techniques. We used Dense Neural Networks (DNNs) and Convolution Neural Networks (CNNs) of different architectures, chosen to limit the model complexity and size. The models were trained during 200 epochs using a mini-batch SGD of size 32, and a dropout rate of 0.50 was employed. Table \ref{fig:centralREALWORLD} shows the performance of the models trained in a centralized approach on the REALWORLD dataset. All evaluations were performed on the global test set. 

\begin{table}[!htb]
\centering
\caption{Centralized learning performance with several models on the REALWORLD dataset}

\begin{tabular}{|l|c|c|c|}
\hline
Model Architecture  & Acc & F-Score (\%)  \\ 
  \hline
196-16C\_4M\_1024D & \textbf{91.39}  & \textbf{92.48} \\ 
196-16C\_4M\_1024D\_512D & 89.75 & 90.72 \\
1024D\_512D  & 84.59 & 85.90 \\
400D\_100D  & 82.41 & 84.76 \\
\hline
\end{tabular}
\label{fig:centralREALWORLD}
\end{table}

From the experiment, the best model is 196-16C\_4M\_1024D. The shape of this model is 196 filters of 16x1 convolution layer, followed by a 4x1 max pool layer, then 1024 units of the dense layer, and finally the softmax layer with 8 units. This model reaches 92.48\% of F-Score, which is far above the current state-of-the-art on REALWORLD dataset, which was previously 81\% of F-Score \cite{realword}. The experiment shows that our CNN model is well-tuned and superior to the state-of-the-art for this dataset. This experiment shows why we have chosen this model for all the subsequent experiments.

We also trained the model on each of the 15 individual client datasets. In that case, the client models' mean F-Score with this local learning approach on the local test-sets was 96.04\%. This result is high but highly biased. If the local models are tested on the global test-set (the combined test-sets of all the clients), the obtained mean F-Score is 51.94\%. Hence a model learned on local data is highly personalized but clearly lacks generalization capability (i.e., this is a classical case of over-fitting)

\subsection{Federated Learning}

The four FL algorithms were run using the CNN model architecture defined in the previous section. Before learning, all models were randomly initialized. We experimented with each FL algorithm during 200 communication rounds. The clients were all set to perform local learning for 5 epochs.  

Table~\ref{fig:resultbl1} presents the results of the learning. The Centralized Server Global accuracy (92.48\%) and Local Personalization accuracy results (96.04\%) give the upper limit in terms of generalization and personalization. Among the FL algorithms, FedDist outperforms all other algorithms for the three measures (Global, Personalization, and Generalization accuracy). It exhibits the best trade-off between generalization and personalization (Generalization accuracy = 74.23\%). FedAvg is the second-best algorithm (Generalization accuracy = 72.99\%) while FedMA (60.09\%) and FedPer (53.01\%) show a trend towards overfitting but without beating FedDist or FedAvg in terms of personalization. Apart from FedMA, all algorithms seem slow in converging before the 200 communication rounds. We detail the results for each algorithm below. 

\begin{table}[!b]

\centering
\caption{Overall results of the learning experiments on the REALWORLD dataset}

\resizebox{\columnwidth}{!}{
\begin{tabular}{|c|c|c|c|c|c|}
\hline

& \multicolumn{5}{c|}{ REALWORLD } \\
\hline

& Global&Personalization& Generalization& Server  & Client \\ 
& F-Score (\%) & F-Score (\%) & F-Score (\%)  & Best Rnd  & Best Rnd\\ 

\hline

Centralized & \textbf{92.48} & N/A & N/A  & 197 (Epoch) & N/A  \\
Local & N/A & \textbf{96.04 $\pm$ 1.77 }& 51.94 $\pm$ 3.39 & N/A & 198 (Epoch) \\
FedAvg & 83.44 & 95.82 $\pm$ 1.53 & 72.99 $\pm$ 1.84 & 198 & 187  \\
FedPer & N/A & 95.46 $\pm$ 1.62 & 53.01 $\pm$ 2.93 & N/A & 190 \\
FedMA & 78.67 & 93.65 $\pm$ 2.21 & 60.09 $\pm$ 1.73  & \textbf{137} & \textbf{171}  \\
FedDist & 84.52 & 95.84 $\pm$ 1.59 & \textbf{74.23 $\pm$ 2.29} & 196 & 191  \\
FedAvg$_{FedDist\_size}$  & 83.97 & 95.74  $\pm$ 2.35 & 73.73 $\pm$ 1.54 & 197 & 183  \\
\hline
\end{tabular}
}
\label{fig:resultbl1}
\end{table}

\begin{figure}[!bt]
\centering
\includegraphics[scale=0.50]{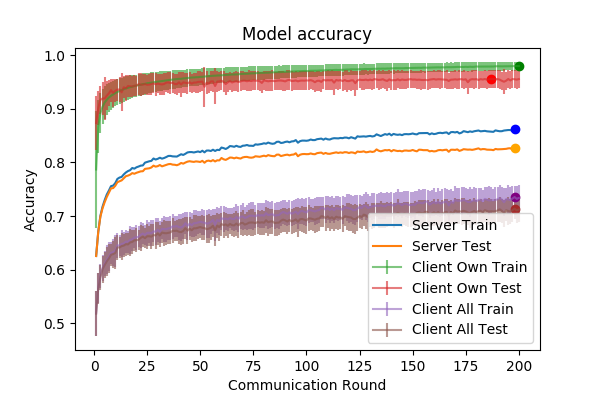}
\caption{FedAvg learning over 200 communication rounds}
\label{fig:fedAvgAcc}
\end{figure}

The FedAvg algorithm, despite its simplicity, seems difficult to beat. As shown in Figure \ref{fig:fedAvgAcc}, with this method, the model at the server level generalizes well on the global test-set with an F-Score of 83.44\% accuracy (which is still far from the centralized learning approach of 92.48\%). The client model obtained a mean F-Score of 95.82\% on the local-test set and 72.99\% on the global test set. The learning curve showed in Figure~\ref{fig:fedAvgAcc} exhibits a classic shape where the learning starts with a steep improvement until it reaches a slow monotonic increase after communication round 20. It seems that the performance on the global-test set would be able to grow with more communication rounds further. We highlight that the slower convergence, compared to the traditional learning approach,  is due to the averaging property of FedAvg, where it produces effects similar to regularization techniques where we limit over-fitting at the cost of slower convergence. The client level learning is less monotonic with sometimes sharp changes in the standard deviation for the Personalization and Generalization accuracy evaluation. This trait can be attributed due to some clients' peculiarities.

\begin{figure}[!bt]
\centering
\includegraphics[scale=0.42]{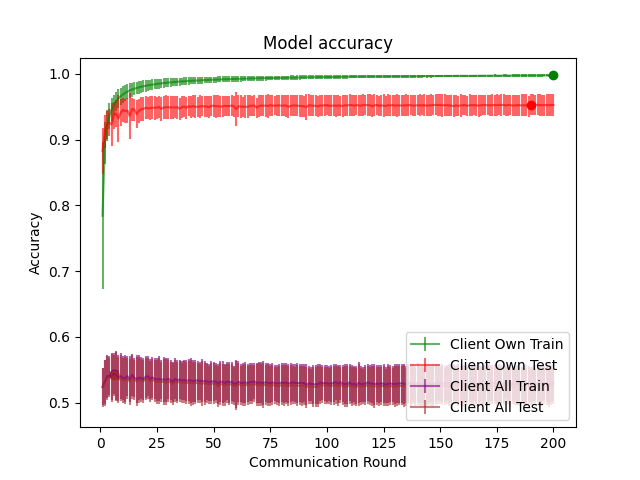}
\caption{FedPer learning over 200 communication rounds}
\label{fig:fedPerAcc}
\end{figure}

With the FedPer algorithm, only the CNN layer part is communicated to the server while the dense layer (i.e., the personalized layer) is kept local. This scheme might explain why the FedPer algorithm does not succeed in generalizing (Generalization accuracy = 53.01\%), as it can be seen in Figure \ref{fig:fedPerAcc}. The personalization layer stays too strong, and such the model is not able to react appropriately to new data. However, the results show that even the personalization feature of FedPer is not significantly better than any of the other FL algorithms.

\begin{figure}[!bt]
\centering
\includegraphics[scale=0.44]{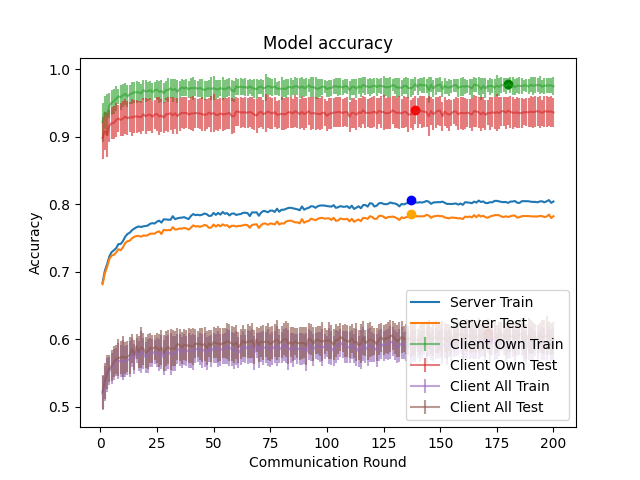}
\caption{FedMA learning over 200 communication rounds}
\label{fig:fedMAAC}
\end{figure}

The FedMA algorithm reached an F-Score of 78.67\% for the Global accuracy, as exhibited in Figure \ref{fig:fedMAAC}. It has the lowest ability to personalize with some minor drawbacks (Personalization accuracy = 93.65\%) while performing moderately on the global test-set (Generalization accuracy = 60.09\%). Furthermore, FedMA has a much higher training cost. Indeed, in our experimentation, 5 local epochs are used to train client models. For FedMA, this means a total of 25 local epochs for each communication round (5 local epochs for each of the first and second layer, and another 15 for the softmax layers).

\begin{figure}[!bt]
\centering
\includegraphics[scale=0.48]{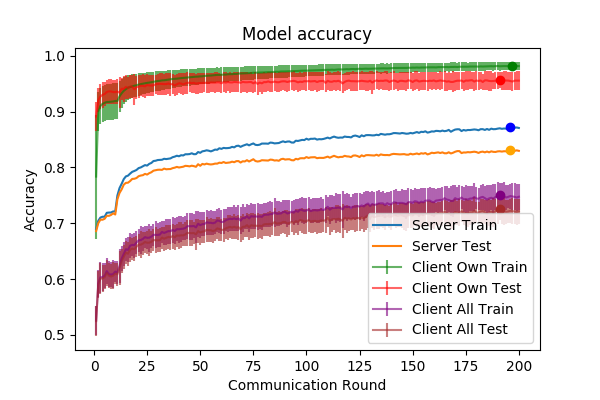}
\caption{FedDist learning over 200 communication rounds}
\label{fig:fedDistAC}
\end{figure}

FedDist presents the best performance overall in our experiments. The learning curve shown in Figure~\ref{fig:fedDistAC} exhibits a two-step behavior. This is because at a certain point, after many communication rounds, no new neuron or filter is added to the global model, and thus, training can stabilize better onward. This behavior is further enforced with a penalty function that takes the current communication round as input to raise new neurons' acceptance rate in later rounds. After no new neurons are needed, we expect the late stages of training similar to the FedAvg algorithm. However, the initial step was sufficient to present better performances than FedAvg. In the end, FedDist gets the best Generalization accuracy (e.g., the best trade-off between generalization and personalization) and could even reach more with further training.

The final global model shape, when FedDist stopped adding new filters and neurons, is 222-16C\_4M\_2250D. From the original size of the CNN, 26 new filters have been added to the convolutional layer, and a total of 1222 neuron units were gained on the dense layer. With the new model size, we then ran another experiment with FedAvg with the same setting (FedAvg$_{FedDist\_size}$ in table {\ref{fig:resultbl1}}) and found that the results were short of the original FedDist implementation. Additionally, when compared to the FedAvg approach with the original model size, some benefits were gained in the global and server accuracy at the cost of a slight loss in personalization.

\begin{table}[!b]
 \centering
  \caption{FedAvg and FedDist results of the learning experiments on the UCI dataset}

 \resizebox{\columnwidth}{!}{
 \begin{tabular}{|c|c|c|c|c|c|}
 \hline

 & \multicolumn{5}{c|}{ UCI } \\
 \hline

 & Global&Personalization& Generalization& Server  & Client \\ 
 & F-Score (\%) & F-Score (\%) & F-Score (\%)  & Best Rnd  & Best Rnd\\ 

 \hline
 FedAvg & 96.96 & 96.73 $\pm$ 1.32 & 96.85 $\pm$ 0.21 & 194 & 198  \\
 FedDist & 96.96 & 96.73 $\pm$ 1.32 & 96.85 $\pm$ 0.21 & 194 & 198  \\
 \hline
 \end{tabular}
 }
 \label{fig:resultbl2}
 \end{table}

Finally, FedDist and FedAvg were also evaluated on the UCI dataset partitioned into 5 uniformly distributed clients. As shown in table {\ref{fig:resultbl2}}, both algorithms give the same results (96.96\% global F-score) in-line with the state of the art (cf. Table~{\ref{table:UCISOTA}}). This outcome can be explained by the fact that FedDist did not add any neuron and thus behaved like FedAvg. This is not surprising since UCI is a very uniform dataset. Nevertheless, this experiment shows that FedDist is versatile enough to adapt to both uniform and heterogeneous datasets.

\section{Conclusion and further work}\label{sec:conclu}
Federated Learning (FL) exhibits clear theoretical advantages over classical centralized learning from a pervasive computing perspective. It provides a solution for distributed learning and, to some extent, privacy preservation. However, little is known about the behavior of such a learning approach and how to evaluate it in a realistic pervasive computing situation where devices can be very mobile and highly specific. Up to now, most of the studies about FL have been conducted in the computer vision area, without considerations for the specific needs of pervasive applications. Furthermore, despite a recent and active research effort, as our paper reveals, the standard simple FL algorithm FedAvg is a difficult approach to beat, and more subtle but complex algorithms do not demonstrate a clear superiority. In this paper, our new FL algorithm FedDist unites the efficiency of FedAvg with the flexibility to make the machine learning model evolve over communication rounds. We also introduce a straightforward methodology to evaluate state-of-the-art FL algorithms and FedDist in the context of a HAR from smartphone sensors.

Our evaluation method showed that FedDist clearly outperformed the other FL algorithms on the three measures of generalization and personalization. Indeed, FL should lead to a high degree of adaptation to the device (high client accuracy on its own data) while keeping a high degree of generalization (e.g., prevent over-fitting, high client accuracy on global data). FedAvg also exhibited such behavior, but more complex FL algorithms such as FedMA and FedPer were not able to keep a high degree of generalization and did not show the superior capacity of personalization. One advantage of FedDist over other algorithms is its ability to make the initial CNN model evolves along with communication rounds. Since deciding on the initial size of a NN model is an open problem, the property FedDist computing the number of new specialized neurons to add automatically to a model provides a flexible way to adapt the model architecture to the task. While FedDist is more computationally intensive than FedAvg, it is far less complex than the FedMA algorithm and with better performance on HAR.

Although these results add credence to the interest of federated learning for pervasive computing, a lot of challenges still remain. The study must be replicated with more datasets and different tasks\cite{BRENON201892}. We also plan to study the robustness of FL in scenarios such as asynchronous learning (devices come and go), a sudden change in client data, communication issues, heterogeneous population of devices (e.g., traveling device), and mismatches between server data and clients (noisy acquisition). Furthermore, long term studies are needed to optimize communication schedule, and life-long learning effects such as catastrophic forgetting \cite{chen2018lifelong}. We also suggest the community set up benchmarks for comparison and replication of research in this area, and we believe that the study presented here is a stepping stone in this direction.
\newpage
\bibliographystyle{ieeetr} 
\bibliography{bibfile.bib}

\end{document}